# Context-Driven Data Mining through Bias Removal and Data Incompleteness Mitigation

Feras A. Batarseh*  
fbatarse@gmu.edu

Ajay Kulkarni*  
akulkar8@gmu.edu

**Abstract**

The results of data mining endeavors are majorly driven by data quality. Throughout these deployments, serious show-stopper problems are still unresolved, such as: data collection ambiguities, data imbalance, hidden biases in data, the lack of domain information, and data incompleteness. This paper is based on the premise that context can aid in mitigating these issues. In a traditional data science lifecycle, context is not considered. Context-driven Data Science Lifecycle (C-DSL); the main contribution of this paper, is developed to address these challenges. Two case studies (using datasets from sports events) are developed to test C-DSL. Results from both case studies are evaluated using common data mining metrics such as: coefficient of determination ($R^2$ value) and confusion matrices. The work presented in this paper aims to re-define the lifecycle and introduce tangible improvements to its outcomes.

## 1. Introduction and Motivation

Historically, most research in AI has been focused on improving the algorithm. In the last decade or so however, the focus has shifted to data - *big data*. Ample amounts of data reshaped AI and renewed its promise and premise. As more machine learning models are deployed across multiple domains [1, 2], new challenges are rising. For instance, the relevance, data types, data quality, and completeness of inputs to a model (dependent variables), effect the significance and 'goodness' of the outputs (independent variables). But how can that be optimized? In the presented method, context is defined and injected into the process to obtain insights that are more relevant and domain-specific. However, in most cases, it is highly challenging to define what *context* is. Context is *infinite* [3], and so data that could be collected to define a complete context is also potentially infinite. For instance, in the sports studies presented, there is an *infinite* amount of information that could be collected and used for contextual awareness. For example, context can consist of data about the weather on the day of the competition, or the type of car that the athlete owns, or their country's birth rate, or the type of shoes worn by them during the competition, or whether the athlete had eggs or cereal for breakfast that day! The point is, the amount and variety of data that could be collected to define the context of the event under study is *infinite*, which makes the scope of this challenge very difficult to capture.

In data collection, and given that any data could be collected (theoretically), then the four Vs of big data (velocity, variety, veracity, and volume) are not representative of the real challenge within the lifecycle of data science; but the main (or first) challenge to be addressed is: what data should be collected *for the problem at hand*? In the studies presented in this manuscript, multiple categorical data columns, coefficients, and correlations are evaluated to define a *context*, multiple approaches are explored, and the results are evaluated statistically and by comparing them to actual results.

The major challenge found throughout the process was the quality of the data (outliers, bias, and incompleteness). As Niels Bohr famously stated: "Prediction is very difficult, especially if it's about the future". The challenge exacerbates however, when the future prediction is an *outlier*. For instance, winning a gold medal or a medal at all is an outlier, very few athletes win medals at the Olympics - one per sport. Same thing applies for most sports events, there is only one winner of the super bowl, one winner of the World Cup, and that winner is the outlier. Contrary to that, if an athlete is historically a winner of medals, for that athlete, not winning a medal becomes an outlier (not the contrary). Therefore, locating outliers depends on the scope, and the subset of the universal dataset that is used. Adding more data to help define context is also dependent on the scope, goals, and the information available in the dataset. Even if we are looking at the same problem, same machine learning model, the slicing and dicing of data is constantly effecting what context consists of. Therefore, if context is that dynamic, then how can it be captured in a data science lifecycle? This paper examines that notion and provides solutions to it using a Context-driven Data Science Lifecycle (C-DSL). The paper is organized as follows: next section discusses the literature review for context, data bias, and data incompleteness. Afterwards, C-DSL is introduced along with the two experimental studies, and in the final section, conclusions and future research plans are presented.

---

*College of Science, George Mason University, 4400 University Dr., Fairfax, Virginia, USA 22030

## 2. Related Works in Contextual Management

As discussed prior, context plays a pivotal role in decision making as it can change the meaning of concepts present in a dataset. The context within a dataset can be extracted and represented as features [4]. Features in general fall into three categories: primary features, irrelevant features, and contextual features. Primary features are the traditional ones which are pertinent to a particular domain. Irrelevant features are features which are not helpful and can be safely removed, while contextual features are the ones to pay attention to. That categorization helps in eliminating irrelevant data but doesn't help in clearly defining context. Another promising method that aimed to solve this challenge, is called the *Recognition and Exploitation of Contextual Clues via Incremental Meta-Learning* [5], which is a two-level learning model in which a Bayesian classifier is used for context classification, and meta algorithms are used to detect contextual changes.

Another method: *context-sensitive feature selection* [6] described a process that out performs traditional feature selection such as forward sequential selection and backward sequential selection. Dominogos's method uses a clustering-based approach to select locally-relevant features. Additionally, [7] introduced a two-tier contextual classification adjustment method called POISEDON. The first tier captures the basic properties of context, and the second tier captures property modifications and context dependencies. Context injections however, have been more successful when they are applied to specific domains. For example, adding context to data has significantly improved the accuracy of algorithms for solving Natural Language Processing (NLP) problems. [8] added context to correct wrongly tagged words. In their paper, the authors have combined the output from the classifier with a set of words manually labeled with context. A transformation-based learning algorithm was used to generate new rules for the classifier. The authors claimed that this method increased the contextual accuracy of their application by 4.8%.

Another example used context for software testing. Context-Driven Testing (CDT), utilizes context to reduce the number of test cases and improve on the validation and verification of software systems. The authors of the paper reported very significant improvements in time and quality of testing results due to context [9].

The issue of deriving context from data however, is even more challenging, for instance, [10] pointed out that data science algorithms without realizing their context could have an *opacity problem*. This can cause models to be racist or sexist (for example). It is often observed that Google translator refers to women as 'he said' or 'he wrote' when translating from Spanish to English. This finding was also verified by Google Inc. Another opacity example is a word embedding algorithm which classifies European names as pleasant and African American names as unpleasant [11]. If a reductionist approach is considered, adding or removing data can surely redefine context, it is observed however, that most real-world data science projects use incomplete data [12, 13]. Data incompleteness occurs within one of the following categorizations: 1) Missing Completely at Random (MCAR), 2) Missing at Random (MAR), and 3) Missing not at Random (MNAR). MAR depends on the observed data, but not on un-observed data while MCAR depends neither on observed data nor unobserved data [14, 15]. There are various methods to handle missing data issues which includes list-wise or pairwise detections, multiple imputation, mean/ median/ mode imputation, regression imputation, as well as learning without handling missing data [12].

All the aforementioned works were challenged with the quality of the data. For example, several types of bias can occur in any phase of the data science lifecycle or while extracting context. Bias can begin during data collection, data cleaning, modeling, or any other phase. Biases which arise in the data are independent of the sample size or statistical significance, and they can directly affect the context of the results or the model. They also affect the association between variables, and in extreme cases, they can even reflect the opposite of a true association or correlation [16].

Based on reviewing multiple works in data science, the most commonly observed bias is *class imbalance* due to *covariate shifts*. Class imbalance is represented by the unequal ratio of categories which can occur due to changes in the distribution of data (covariate shifts). Class imbalance depends on four factors: 1) degree of class imbalance 2) the complexity of the concept represented by the data 3) the overall size of the training size and 4) the type of classifier [17]. Datasets with imbalance create difficulties in information retrieval, filtering tasks, and knowledge representation [18, 19].

In this paper, *context* is extracted by deploying a variety of statistical methods: *data imputation*, creation of a *generic coefficient*, adding data columns (such as: host country, sport, GDP, height, weight, and age), *weighted modeling*, and *mitigation of bias*. The details about the method (main contribution of this paper) and techniques used are presented in the next section.

## 3. Context-Driven Data Science Lifecycle

C-DSL has five main steps (Figure 1). Those five steps are represented in two experiments (Olympics medal predictions and the UEFA Champions League winners and losers). In the first step, data cleaning and wrangling are performed. In the literature [22, 23, 24] it is indicated that data cleaning helps to build robust and more reliable models. Data wrangling is considered one of the most expensive phases in the data science lifecycle. During that phase, multiple decisions are taken, that includes: eliminating subsets of data, filtering, and aggregation. In the second step of C-DSL, context is injected. For experiment 1, that is done by adding details like year, host city, sport, name of athlete, country of the athlete, medal type (gold, silver, and bronze) and athlete's demographical data.

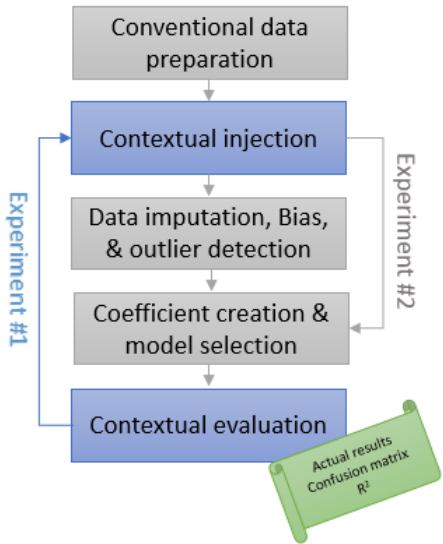

Figure 1: C-DSL

For experiment 2, context is injected by collecting, cleaning and generating sentiment scores from social media text (tweets). For step 3, Data imputation, bias removal, and outlier detection are performed for the first experiment (explained in great details in the next section). In the fourth step of C-DSL, prediction models are built for experiment 1, while a coefficient is created for experiment 2 and used for predictions. In the final step of C-DSL context is evaluated using confusion matrices, and model quality measure such as R-squared; and performance of the models is compared with actual results of the sports events. C-DSL is meant with the continuous fine-tuning of data until a certain 'contextual' sweet spot is achieved. The proposed combination of statistical methods are tools that are used to reach that contextual understanding of the dataset, and be able to then predict based on that.

In the Olympics experiment, outliers and bias in data lead to results that are barely better than the conventional process, but in the second experiment (Champions League), and after understanding context due to data imputation and inference, a coefficient is proven very successful in predicting the results of a tournament with very high accuracy. In the next section, an in-depth explanation of the implementation of C-DSL for both experiments is presented.

## 4. Experimental Work

This section aims to test and evaluate the method presented in this paper, and present the detailed process followed to define it.

### 4.1 Experiment #1 (Olympics Predictions): Data Preparation and Statistical Deployments

In this experiment, an application of sports predictions has been developed using summer Olympics data between years 1896 and 2016. Two datasets are pulled from Kaggle.com. The first dataset has 31,165 observations, and the second dataset consists of more than 200,000 observations. The datasets could be requested from the authors. In the conventional data preparation step, winter data is filtered out (the aim is to predict next summer Olympics medal counts by country and sport). Summer data is then checked for missing values. Information on some athletes was missing, such as: Age, Height, and Weight. A function from the R "`mice`" package "`md.pattern()`" is used for getting insights into the patterns of missing data. Additionally, it is for example observed that 1,888,464 athletes didn't win any medals; that is represented by *nulls* in the medals' column. Nulls are then replaced by "No medal", because some models in R choke when dealing with null values. The missing values (count: 114,900) are then imputed using the ***Multivariate Imputation by Chained Equation*** (MICE) technique [20]. After that, columns such as Sport, Gender, Age, Height, and Weight are used as *context*. This operation is performed by Predictive Mean Matching (PMM) method in R using the "`mice`()" function. *Fifty* iterations of imputations were required to create all the missing data - approximately 15 hours to complete the entire process.

Outlier detection is then performed, using **Local Outlier Factor** (LOF). LOF It is a density-based outlier detection technique [21]. The main reason for choosing this method is the type of variables in the dataset. In outlier detection it is essential to convert categorical variables into numerical variables. In addition to that the numerical variables are scaled using the "`scale()`" function. Initially, there are 5 columns (Sport, Gender, Age, Height, and Weight) in the data but after performing scaling and encoding of values in categories, fifty three representative columns are created (as iterative combinations of these columns). The function "`lofactor()`" is used with "k = 5" for outlier detection. In the function, k denotes the number of nearest neighbors that represent the locality used for estimating the density.

Afterwards, model selection was deployed; regression and random forests are used for this experiment. In the first part, a simple linear regression model is built in R using the `lm()` function. Further, predictions per sport per country are developed using multiple linear regression. For that purpose, six different weight scenarios are used, and the models are tweaked to enforce more significance on recent years. For random forests, classification is based on the type of the medal (gold, silver, bronze, and no medal), Sport, Gender, Age, Height, and Weight of the athlete. To perform the classification, medals are encoded by numbers ("Gold = 1", "Silver=2", "Bronze=3" and "No medal=4"), and then the model is trained on the entire dataset from

1896 to 2012 (using "`randomForest`" and "`ranger`" packages in R). The results of this experiment were not very convincing (presented in experimental results), although much better than conventional predictions. This experiment reflected the importance of tuning the value of k, creating a coefficient, and the criticality of inference, something that is deployed in the second experiment.

### 4.2 Experiment #2 (Text Mining for Context): Setup and Coefficient Creation

In this experiment, social media data are collected to be the main driver for Context. In sports, it is safe to assume that the fans of a sports team can reflect or influence the team's status, and maybe even help in predicting the outcomes of that team. This study calculates sentiment scores for text relevant to the Champions League (a European Clubs Soccer Championship), and uses that as the context of a team to help predict whether the team will perform well in next stages or not. The sentiment score for each post or tweet is normalized on a -7 to +13 scale. The R "`tm`" package is used to scan through the tweets and assign scores based on a set of predefined words. Once all the tweets have scores, a *coefficient* is created: Average Team Sentiment Score (ATSS). It is defined as: (Sum of Sentiment score of all tweets at the team level) / (Count of tweets at the team level).

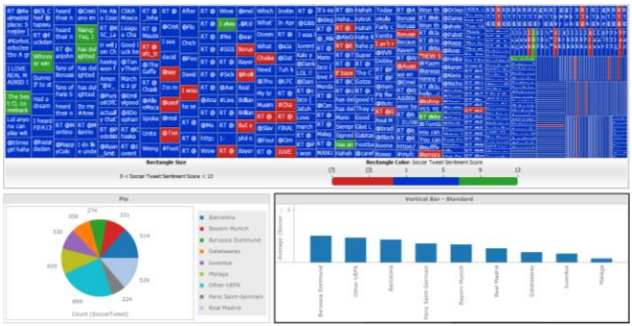

Figure 2: Sentiments of tweets and counts of tweets per team

The idea of the coefficient is to represent the team's popularity and the sentiments of its fans. This study was deployed for eight teams: Barcelona, Real Madrid, Juventus, Bayern Munich, Borussia Dortmund, Galatasaray, and Paris Saint Germain. Figure 2 shows a data visualization that illustrates the results of sentiments tweets. It shows a sample of all tweets and their sentiment values. Red is a negative sentiment, green is a positive sentiment, and blue is neutral. The main takeaway from Figure 2 is to visualize the distribution of sentiments from the tweets on all the different teams. It can be observed from the heat map that most of the sentiments are neutral (blue), while the pie chart indicates that Barcelona F.C. has the highest number tweets.

Additionally, Figure 3 shows the sentiments when aggregated to the country level. For example, tweets from China and Russia about the tournament are negative on average, and ones from USA and Canada are positive on average, while Europe varies. The results for both experiments 1 and 2 are presented in the following sub-section.

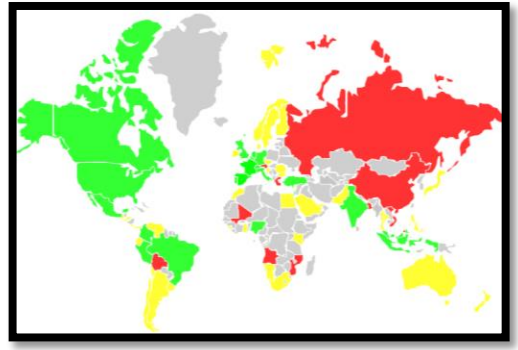

Figure 3: Sentiment score heat map by country

### 4.3 Experimental Results: Olympics Predictions

After deploying C-DSL steps, the predictions for the first experiment were acceptable, certainly better than without deploying context, however, not very satisfactory. The bar plot in Figure 4 the actual number of medals (blue bar on the left) and orange color (on the right) indicates predicted number of medals through C-DSL.

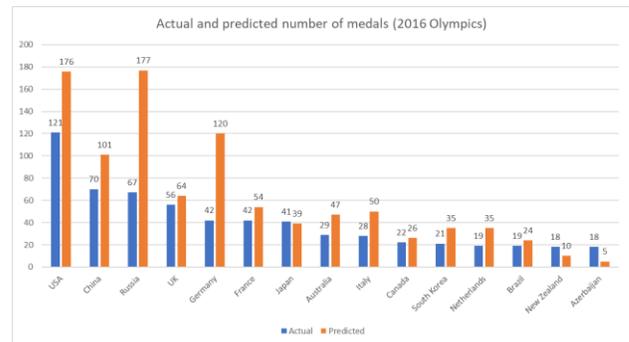

Figure 4: Actual and predicted number of medals

The observed adjusted $R^2$ value for the simple linear regression model is 0.5488. It can be easily observed that for Japan, Canada, Brazil, New Zealand, and the UK the actual number of medals and predicted number of medals are very close, and potentially useful for decision making.

In the second round, after applying *weights* for predicting number of medals per sport, for top 5 countries, it is observed that all the models are predicting better number of medals for: USA, China, Russia, and Germany, and that is reflective of actual results. In the case of the UK, all the models were close to the actual number of medals (90% accuracy). In Table 1, the best results from C-DSL are presented. Results from C-DSL are much better than the conventional regression process. Furthermore, Table 2 shows results compared to actual events (confusion matrix). The model is able to predict 13 correct records for (1 Gold), 3 correct records for (2 Silver) and 9 correct records for (3 Bronze).

| Country | Sport | Actual | Conventional | C-DSL |
|---|---|---|---|---|
| USA | Gymnastics | 12 | 18 | 14 |
| UK | Gymnastics | 7 | 11 | 7 |
| UK | kayaking | 4 | 6 | 5 |
| UK | Athletics | 7 | 8 | 6 |
| UK | Sailing | 3 | 5 | 4 |
| UK | Boxing | 3 | 4 | 3 |
| UK | Taekwondo | 3 | 3 | 2 |
| UK | Triathlon | 3 | 3 | 2 |
| UK | Tennis | 1 | 4 | 2 |
| UK | Shooting | 2 | 5 | 2 |
| China | Table Tennis | 6 | 5 | 6 |
| China | Athletics | 6 | 8 | 7 |
| China | Taekwondo | 2 | 3 | 3 |
| China | Boxing | 4 | 4 | 4 |
| Russia | Wrestling | 9 | 9 | 9 |
| Germany | Kayaking | 7 | 7 | 7 |
| Germany | Shooting | 4 | 6 | 5 |
| Germany | Equestrian | 6 | 7 | 8 |

Table 1: Selected results for different sports for top 5 countries

|  |  | Reference/Actual | | | |
|---|---|---|---|---|---|
|  |  | 1 | 2 | 3 | 4 |
| **Prediction** | 1 | **13** | 6 | 6 | 73 |
|  | 2 | 9 | **3** | 10 | 61 |
|  | 3 | 5 | 12 | **9** | 63 |
|  | 4 | 638 | 634 | 678 | **11468** |

Table 2: Confusion matrix for predictions

The overall accuracy of the random forests model is 83.96%, which usually reflects high accuracy, however, due to data imbalance (which could be also considered as an outlier issue), the results in Table 2 are potentially a result of a model that is underfitting. The claim made in this scenario is that context can be used as a pointer to such unclear data lifecycle dilemmas.

### 4.4 Experimental Results: Text Mining for Context

After calculating the sentiments and the activities for all tweets, an aggregation of ATSS (the coefficient) for every team is created. The coefficient reflects the ATSS for every team, as well as the count of tweets per team (i.e. interest and hype surrounding that team). The results from this experiment are very successful (more than Experiment 1). When the coefficient-by-team is sorted (as Figure 5 shows), the highest two teams are the teams that reached the final game in that tournament. Followed by the other two semi-finalists, and then followed by teams in the quarter finals, that result indicates how contextual awareness of the tournament (through data from fans for instance), can provide predictions with high statistical confidence.

The predictions for this study are much more indicative of actual events than when compared to the UEFA ranking of those teams for instance, or expectations based on stars playing for them, or any other conventional method. It is important to note however that these results are not tested across multiple types of tournaments, rather only for one year (2013). That is due to the availability of the data, this work however is certainly ongoing, and we aim to deploy the same method for multiple tournaments. In 2013, Bayern Munich won the tournament, and teams such as Barcelona and Paris Saint Germain unexpectedly lost.

**Team coefficient reflecting the winner**

| Team | TeamCoeffecient2013 |
|---|---|
| Bayern Munich | 36.585 |
| Borussia Dortmund | 33.585 |
| Real Madrid | 29.542 |
| Barcelona | 27.542 |
| Paris Saint Germain | 27.35 |
| Juventus | 25.883 |
| Malaga | 25.542 |
| Galatasaray | 24.04 |

Figure 5: Team coefficient very indicative of actual results

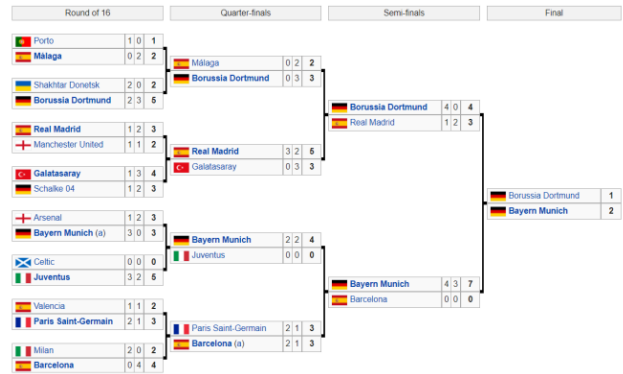

Figure 6: Actual results of 2012-13 UEFA Champions League [25]

C-DSL, based on contextual understanding of the fans, the hype, social media attention, and collective knowledge is able to predict the winner. The work presented in both experiments has potential for improvements, and is still undergoing, conclusions and next steps are presented in the next section.

### 5. Conclusions and Next Steps

In this paper, a Context-driven Data Science Lifecycle (C-DSL) is introduced and tested for applications of sport predictions. It can be concluded from the results that con-text plays a crucial role for prediction. In addition to that, based on our experiments, techniques for data imputation, bias, and outlier detection have a significant influence in C-DSL. Two experiments are performed, they utilize C-DSL steps slightly differently, and they have different success rates. However, both experiments are successful in providing better outcomes than the conventional data science lifecycle. The method presented in this paper is deemed to be very specific to certain types of data sets, and certain data mining problems. The experiments presented illustrate it as a punctual solution to a broad problem, however, C-DSL could be generalized to many other types of data sets. For future steps, we aim to do the following: 1. Develop a tool that automates the process of C-DSL, 2. Experiment with more types of sports events, 3. Redefine C-DSL to create a more unified and generic process that applies to all types of datasets, 4. Identify other data sets that have a variety of data types and test them through C-DSL, 5. Deploy C-DSL for upcoming summer sports tournaments and compare the results to media and experts predictions.